# Geração Automática de Painéis de Controle para Análise de Mobilidade Urbana Utilizando Redes Complexas

**Victor Dantas, Henrique Santos, Carlos Caminha, Vasco Furtado[1]**

[1]IPPGIA - Programa de Pós-Graduação em Informática Aplicada - Universidade de Fortaleza - Fortaleza, CE - Brazil

`{victordantas2,hos}@edu.unifor.br, {caminha,vasco}@unifor.br`

***Abstract.*** *In this paper we describe an automatic generator to support the data scientist to construct, in a user-friendly way, dashboards from data represented as networks. The generator called SBINet (Semantic for Business Intelligence from Networks) has a semantic layer that, through ontologies, describes the data that represents a network as well as the possible metrics to be calculated in the network. Thus, with SBINet, the stages of the dashboard constructing process that uses complex network metrics are facilitated and can be done by users who do not necessarily know about complex networks.*

***Resumo.*** *Neste artigo descrevemos um gerador automático para apoiar o cientista de dados a construir, de forma amigável, painéis de controle a partir de dados que são representados como redes. O gerador chamado SBINet (Semantic for Business Intelligence from Networks), possui uma camada semântica que, via ontologias, descreve os dados que representam uma rede além das métricas possíveis de serem calculadas na rede. Agindo assim, com SBINet, as etapas dos processos de construção de uma rede complexa, cálculo das métricas e criação de objetos de interação são facilitadas e podem ser feitas por usuários que não necessariamente conhecem de redes complexas.*

## 1. Motivação e Problema

As ferramentas de Business Intelligence - BI - se destacaram pela capacidade de colaborar nos processos de análise e compreensão de grandes volumes de dados. Painéis de controle (*dashboards*) são as plataformas típicas usadas para esse fim e há uma grande variedade de ferramentas (Qlik Sense, Pentaho, Tableau, Microstrategy, entre outros) no mercado para auxiliar sua construção por cientistas de dados não necessariamente especialistas em programação.

A despeito desse avanço, pouco foi feito nessas plataformas para apoiar ao usuário não-programador a construção de painéis de controle a partir de redes complexas. Nesse contexto particular, há ainda o problema de que analistas de dados podem não dominar os conceitos e métodos que permeiam o contexto das redes complexas de forma a ser capaz de aplicá-los no processo de análise de dados. As ferramentas tradicionais de manuseio de redes complexas como Gephi, Pajek, e Hive Plots pressupõem o conhecimento teórico do usuário, de forma que o mesmo seja capaz de interpretar a semântica dos dados que compõem a rede e as métricas que delas podem ser extraídas.

Nossa pesquisa se insere nesse contexto, propondo uma ferramenta que apoie o cientista de dados a construir, de forma amigável, painéis de controle a partir de dados que são representados como redes. Essa ferramenta, chamada *SBINet* (Semantic for Business Intelligence from Networks), possui uma camada semântica que, através de ontologias, descreve os dados que representam uma rede além das métricas possíveis de serem calculadas para uma dada rede em particular. Agindo assim, com *SBINet*, automatizamos etapas dos processos de construção de uma rede complexa, cálculo das métricas e criação de um BI que permita explorar as métricas e propriedades da rede gerada.

Mostramos a aplicação de *SBINet* no contexto de cidades inteligentes e sua aplicação para apoiar à construção de painéis de controle a partir de duas redes complexas representando o contexto de mobilidade urbana: uma descrevendo o sistema de transporte coletivo e outra descrevendo um sistema de bicicletas compartilhadas.

## 2. Definição da semântica de métricas de redes complexas no contexto de mobilidade urbana

A associação entre as métricas de uma rede de mobilidade urbana e os seus respectivos significados no domínio é uma etapa essencial da abstração das técnicas e métodos de redes complexas na construção de painéis de controle. Na verdade, este é um pré-requisito para a garantia do objetivo de produzir painéis de controle contendo uma interpretação adequada dos resultados.

Neste trabalho foram realizados estudos de caso em duas redes reais: a primeira contendo um sistema de bicicletas compartilhadas e a segunda contendo a infraestrutura da oferta de ônibus. Na rede de bicicletas um vértice representa uma estação, local onde os usuários podem retirar e deixar as bicicletas. As arestas nessa rede representam a retirada de bicicletas, por parte dos usuários, partindo de uma estação de origem para uma estação de destino. Já na rede de ônibus, os vértices representam paradas de ônibus e uma aresta que conecta dois vértices $vi$ e $vj$, que representam a passagem de uma ou mais linhas de ônibus em sequência pelas paradas de ônibus $vi$ e $vj$, em outras palavras elas representam a oferta. Ao longo desta seção será discutido a respeito do significado que as métricas e conceitos de redes complexas têm nos contextos de redes de mobilidade urbana. Essas associações serão abordadas por meio de exemplos de aplicação dos indicadores no contexto dos sistemas supracitados. Essa rede tem sido amplamente estudada em [Caminha et al. 2016], [Caminha et al. 2017] e [Ponte et al. 2016].

Em nossa pesquisa, identificamos que algumas métricas só fazem sentido em redes que representam caminhos. Por esse motivo, esta seção será dividida duas partes. Na primeira, serão detalhados os significados de métricas que se aplicam nas duas redes estudadas. Posteriormente, serão detalhadas métricas que só tem significado na rede de ônibus, devido a mesma representar caminhos que os veículos fazem pela cidade. Diferente da rede de ônibus, a rede de uso de bicicletas compartilhadas não tem caminhos na semântica das ligações de suas arestas[1]. Nessa rede, uma conexão representa apenas que um usuário retirou uma bicicleta de uma estação (vértice) e colocou em outra, por esse motivo avaliamos que métricas de caminhos, como *betweenness* e *excentricidade*, não tem sentido nessa rede.

---

[1] As bicicletas não são rastreadas e não necessariamente seguem rotas predefinidas como é o caso dos ônibus que além de possuírem GPS seguem roteiros predeterminados.

**Associações aplicáveis às duas redes**

A média aritmética do número de conexões que cada uma das estações possui com as demais nos permite medir o seu grau de conectividade médio. No sistema de bicicletas compartilhadas esta medida indica o grau de interligação entre as estações, permitindo por consequência a identificação de estações que possuem um grau de conectividade abaixo, acima e dentro da média. Já na rede de ônibus, a média de conexões indica as paradas com alto grau de conectividade, o que poderia contribuir na identificação de paradas com concentração de oferta ou locais candidatos para a criação de terminais de ônibus.

Esta medida permite entender como se dá a distribuição de conectividade média entre os nós do sistema. Apesar de parecer trivial, esta medição requer cuidados especiais, considerando que redes não necessariamente possuem uma distribuição normal ou linear sobre o número de conexões. Um exemplo disso são as redes que possuem poucos nós com alta conectividade e muitos nós com baixa, o que configura uma distribuição que segue uma lei de potência. A medição do nível de conectividade, a sua distribuição e os picos e vales desta representação podem ser úteis na comparação com outras redes, assim como a identificação de pontos críticos no sistema. Em redes complexas este indicador é obtido por meio do cálculo do *Grau Médio*, que é dado pela média aritmética do número de arestas que cada nó da rede possui conectada a ele mesmo.

No entanto, se as redes forem ponderadas e o pesos das arestas representam a oferta ou procura, a relação entre o volume de demanda e o volume de entrega está diretamente relacionada ao desempenho dos sistemas de mobilidade. O não atendimento da demanda poderia ser interpretado como pontos de saturação do sistema. Os pontos com elevado saldo de oferta, mas de baixa procura indicariam pontos de ociosidade. Medir o quanto um sistema está efetivamente sendo utilizado permite detectar os locais onde a média de utilização está abaixo, acima ou dentro do esperado. Neste modelo os pesos das arestas são capazes de receber diversas interpretações variando conforme altera-se o dado que ele representa, como: números de linhas de ônibus que passam pela aresta, quantidade de usuários que trafegam entre duas paradas de ônibus, número de empresas de transporte que atuam entre dois nós, capacidade máxima que o sistema suporta, número de retiradas e número de devoluções de bicicletas entre duas estações de bicicletas, entre outros.

O nível de utilização de uma estação de bicicletas ou uma parada de ônibus é obtido por meio do cálculo do *Grau Médio Ponderado* dos vértices que as representam. A diferença entre o grau médio e o grau ponderado é que o cálculo leva em conta o peso das arestas e não a quantidade de conexões. Calculamos o *grau médio ponderado* de uma rede pela média aritmética do peso de suas arestas.

É possível que grandes sistemas de mobilidade urbana possam ser vistos como um conjunto de comunidades, estas subdivisões auxiliam no controle, no planejamento e na gestão dos sistemas. Detectar estas comunidades pode ser altamente relevante para a compreensão da rede e de seu funcionamento. Comunidades que caracterizam a oferta de linhas de ônibus, por exemplo, podem ser usadas em comparação com comunidades caracterizando origens e destinos realizados pelos passageiros para identificação de gargalos no sistema como, por exemplo, ônibus lotados [Caminha et al. 2016]. Em sistemas de bicicletas compartilhadas a existência de *Comunidades*, evidenciam em alguns casos a existência de obstáculos físicos, a ausência de infraestrutura adequada ou mesmo elevada distância entre estações ao ponto de tornar inviável ao usuário a utilização de uma rota.

Esta tarefa é realizada por meio do cálculo da *modularidade* da rede, que indica o quanto a mesma é capaz de ser segregada em comunidades. Uma alta Modularidade sugere que a rede é formada por um complexo conjunto de comunidades, sendo a recíproca verdadeira, quanto menor a *modularidade* menor será o número de comunidade que formam esta rede. Na ocorrência de comunidades ou partes menores da rede que não possuem conectividade alguma com as demais passamos a classificar estas áreas como inacessíveis. Esta medida é dada pela contabilização do número de componentes desconectados à rede.

Calcular um indicador que permita avaliar o quanto o sistema é bem conectado em relação ao máximo possível de conexões existentes, também possui uma relevância. Esta métrica permite uma análise sobre o nível de interligação entre as paradas de ônibus ou as estações de um sistema de bicicletas compartilhadas. Em outras palavras, o quão conectados entre si os nós estão. Apesar de útil na comparação entre pequenas redes, existe uma degradação da sua utilidade diretamente proporcional ao tamanho da rede, contudo, esta métrica pode indicar ainda outros fatores que afetam o sistema e que impeçam uma densidade completa. Redes que não possuem arestas repetidas e que não são direcionadas possuem no máximo $v*(v-1)/2$ arestas, onde $v$ é o número total de vértices. Desta forma definimos a *Densidade* $D(G)$ de uma rede $G$ como sendo a razão entre o número de arestas $m$ que a rede possui e o número máximo possível de arestas [Lewis 2009].

Considerando que o número de conexões de uma parada de ônibus ou de uma estação de bicicletas e o peso de suas arestas normalmente variam dentro da própria rede, medir o nível desta possível diferença indica o grau de aleatoriedade em uma rede. Isso significa compreender o quanto pode variar a demanda por ônibus em uma parada ou o quanto pode variar as retiradas ou devoluções em uma estação de bicicletas. A *Entropia* é a métrica que mede a aleatoriedade de uma rede por meio da distribuição dos seus graus a partir da medição da distribuição de probabilidade dos estados em que a rede se encontra.

Por fim, a centralidade de uma parada de ônibus ou uma estação de bicicletas mede o nível de posicionamento destas em relação ao centro do sistema, ou em outras palavras o quanto este vértice é capaz de estar localizado no centro da rede. A *Centralidade de Grau* de um vértice é dada pelo percentual de vértices que estão conectados diretamente a ele em relação a todos os vértices da rede. [Estrada and Rodríguez-Velázquez 2005] formaliza este conceito definindo a *Centralidade de Grau* $C(v)$ de um vértice $v$ como sendo a razão do seu grau $d_v$ e o número total de vértices $v$ da rede menos ele mesmo.

**Associações aplicáveis exclusivamente à rede de ônibus**

O método de identificação do menor caminho para se chegar do nó $v$ ao ponto $v'$ em uma rede pode variar em função do método utilizado, por exemplo: o *caminho mínimo* que liga duas paradas de ônibus sendo possível ser completamente diferente dependendo do critério de medida adotado. Alguns exemplos disso, seria: calcular *caminhos mínimos* em distância, em quantidade de saltos (número de paradas de ônibus percorridas) ou em tempo. Chamamos de *caminho mínimo* de uma rede o menor conjunto de vértices que precisariam ser visitados para chegar de $v$ até $v'$ [Boccaletti 2006].

A identificação das distâncias máximas que um usuário do sistema de ônibus deveria percorrer para chegar de um ponto $v$ do sistema até um ponto $v'$ auxilia na compreensão do grau máximo de distância entre dois pontos da rede. Esta métrica é útil na avaliação da facilidade de acesso ou ausência desta entre pontos do sistema, na qual se equivale ao *diâmetro da rede*. A identificação do nó mais distante $v'$ de um dado nó $v$ de partida, se dá pelo mapeamento individual das conexões mais distantes ou mais onerosas dentro da rede. Em redes complexas chamados de *excentricidade de um nó v*, a maior distância que interliga o nó $v$ e o nó $v'$ mais distante dele dentro da rede. Segundo [Boccaletti 2006], o *Diâmetro* de uma rede é a *excentricidade* máxima do conjunto de todas as *excentricidades* da rede.

[Newman 2003] define a *distância de caminho médio* $l(G)$ de uma rede $G$ como sendo a média de todos os caminhos entre todos os vértices. Matematicamente essa medida é calculada pela razão entre o somatório de todos os *caminhos mínimos* pela quantidade de *caminhos mínimos*. Ao aplicarmos esta métrica sobre uma rede, temos uma forma de avaliar e comparar a facilidade ou ausência desta em se chegar de um ponto a outro da sua estrutura.

Analisar os *caminhos mínimos* entre as arestas de uma rede, mais especificamente calculando sua centralidade de *betweenness*, é útil na compreensão do quanto uma determinada parada de ônibus é importante. Tal centralidade refere-se ao percentual de ocorrência de um dado vértice $v$ em relação a todos os *caminhos mínimos* entre todos os vértices da rede [Freeman 1977].

## 3. Atribuição de significado aos dados e aos indicadores de mobilidade

Em seu uso tradicional, ferramentas de BI recebem como entrada dados tabulares (em formato CSV, bases de dados relacionais ou similares). Dados nesses formatos não possuem uma semântica associada, de forma que seu entendimento e processamento pela ferramenta só são possíveis com o auxílio de um especialista que compreenda o significado de cada coluna presente: é o desenvolvedor do BI que, ao interagir com a ferramenta, determina a coluna que será plotada, que coluna será agrupada e que operação ou cálculo será efetuado, tomando como base as necessidades do usuário. Nesse contexto, fizemos utilização de tecnologias da Web Semântica para permitir que o significado dos dados a serem apresentados para nossa aplicação estivessem explicitados, de forma que uma aplicação inteligente pudesse fazer uso dos dados e construir painéis de controle sem a necessidade de informação extra de um especialista. Para isso, foram desenvolvidas:

1. Uma ontologia de domínio[2] que modela todas a entidades envolvidas e seus relacionamentos, como possíveis significados de nós e arestas de uma rede.
2. Uma ontologia de indicadores[3] que explicita operações a serem realizadas sobre os dados e seus significados.

A Fig. 1 mostra parte da ontologia de domínio relativa aos conceitos envolvidos no sistema de compartilhamento de bicicletas. Pode-se perceber que o conceito central representa uma viagem de bicicleta a qual possui uma estação de origem e outra de destino, bem como um usuário que realiza a viagem. Na Fig. 2 pode-se ver a definição de uma rota de ônibus que tem uma parada de origem e outra de destino, além de um peso associado que representa a quantidade de ônibus que fazem aquela rota.

---
[2] http://hadatac.org/ont/qoe-m#
[3] http://hadatac.org/ont/qoe#

Para o desenvolvimento da ontologia de indicadores, fizemos uso de conceitos utilizados na visualização de dados por ferramentas de BI:

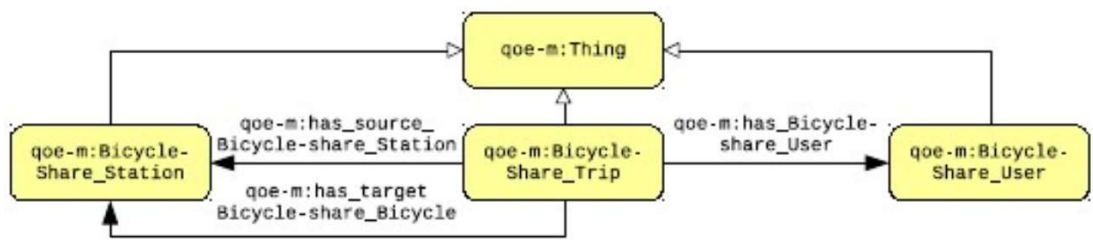

**Figura 1. Ontologia de domínio: compartilhamento de bicicletas**

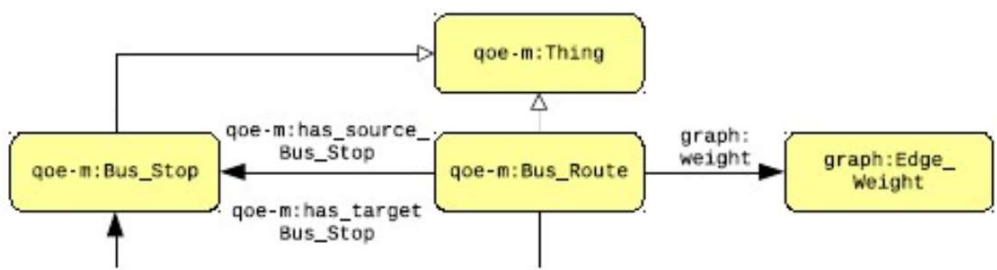

**Figura 2. Ontologia de domínio: rotas de ônibus**

- Dimensão: Um valor de entidade que normalmente não pode ser agregado, geralmente usado como uma linha ou coluna;
- Medida: Um valor de entidade que pode ser utilizado para se calcular algo, por exemplo, uma soma ou média. Geralmente utilizado para exibição ou plotagem.

A ontologia de domínio serve de insumo para a ontologia de indicadores, fazendo uso dos conceitos descritos para descrever as operações de cálculo. Nesse artigo optamos por descrever dois indicadores: "Menor rota entre duas paradas de ônibus" e "Média de interligações entre estações de bicicleta". A Fig. 3 exibe parte da ontologia de indicadores que contempla o indicador de menores rotas. Como visto na seção anterior, esse indicador utiliza algoritmo para cálculo de *caminho mínimo* $minimum\_path(v, v')$. Nesse sentido, definimos como dimensão a aresta ponderada que representa uma rota de ônibus *qoe-m:Bus_Route* e como medida definimos que fosse executada a operação de soma *qoe:Sum* sobre o peso de cada aresta.

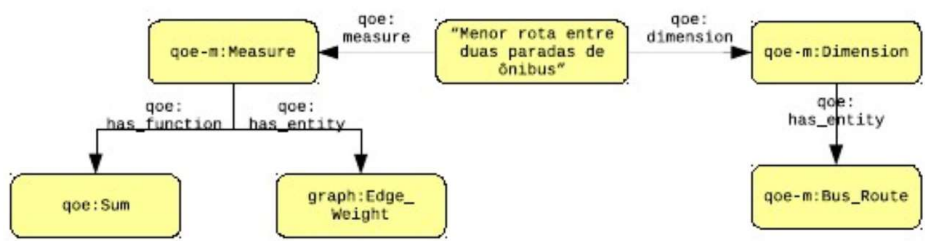

**Figura 3. Ontologia de indicadores: menores rotas**

Já na Fig. 4, mostramos o indicador de média de interligações que é calculado através do algoritmo de grau médio da rede. Para isso, definimos o indicador como possuindo apenas uma medida e nenhuma dimensão. A operação executada é a média *qoe:Average* sobre a quantidade de conexões de uma determinada estação.

Para esse trabalho, fixamos o tipo de visualização para cada indicador definido. Para o de menores rota definimos "mapa", enquanto para a média de interligações definimos "histograma". O processo de anotação dos datasets utilizando as ontologias descritas acima é detalhado em [Santos et al. 2015] e [Santos et al. 2017] e será exemplificado na próxima seção.

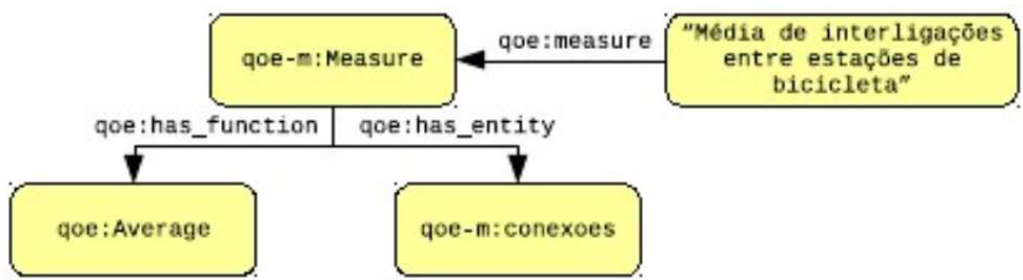

**Figura 4. Ontologia de indicadores: média de interligações entre estações**

## 4. Geração de automática de dashboards

O *SBINet* exige a entrada de dois e somente dois datasets para a execução de todo o processo de construção dos painéis de controle. Eles são inseridos via interface do próprio sistema. O primeiro dataset contém dados dos nós da rede enquanto o segundo as arestas que conectam estes nós. Ambos os datasets foram anotados como descritos em [Santos 2015] e [Santos 2017]. A Fig. 5 ilustra parte destes dois arquivos após o processo de anotação. O dataset da esquerda contém os dados das estações de um sistema de bicicletas compartilhadas enquanto o dataset da direita contém os dados das viagens dos seus usuários, a zona superior de ambos os arquivos contém as anotações semânticas dos dados em Turtle (uma das sintaxes recomendadas pelo W3C para serialização do modelo RDF), inseridas durante o processo de anotação semântica.

**Figura 5. Datasets de estações e bicicletas compartilhadas**

Após a entrada dos datasets, rotinas internas realizam a construção do grafo em modelo RDF a partir das anotações contidas nestes arquivos criando, assim, um grafo de conhecimento sobre a rede. De posse do grafo é possível realizar consultas SPARQL sobre o grafo para que seja possível identificar o domínio dos dados inseridos no sistema. Por exemplo, a consulta abaixo identifica qual o tipo de sistema representado nos arquivos de nós verificando as ontologias as quais os nós estão ligados:

```
select ?type ?typename where {
 ?type a [] .
 bind(if(exists{?type a qoe-m:Bicycle-Share_Station.},
  "type Bicicletas Compartilhadas",
   if(exists{?type a qoe-m:Bus_Stop},"type Ônibus",
    if(exists{?type a qoe-m:Subway_Station},"type Metrô",
     "unknown type" ))) as ?typename )
}
```

Associado ao domínio de um sistema de mobilidade urbana, existe um conjunto de indicadores de redes complexas que podem ser utilizados na sua caracterização, conforme exemplificado no segundo capítulo deste trabalho. Contudo, considerando a diversidade de dados e formatos que os dados podem ter, o *SBINet* precisa identificar quais métricas podem ser calculadas a partir dos dados disponibilizados.

Esta tarefa também é executada por meio de consultas SPARQL que verificam se todas as condições para o cálculo de cada métrica são satisfeitas. Por exemplo, conforme ilustrado na Fig. 4, o cálculo da média de interligações entre as estações de bicicleta exige a informação do número de conexões que cada uma das estações possui, a consulta abaixo é capaz de verificar se esta condição é satisfeita pelos datasets inseridos pelo usuário detectando a presença de campos ligados à ontologia. Este processo é repetido para todas as métricas, permitindo assim a identificação de quais métricas poderão ser calculadas.

```
SELECT ?connections WHERE{ ?connections a qoe-m:conexoes.}
```

O processo de cálculo das métricas não requer intervenção por parte do usuário, uma vez que os datasets foram devidamente anotados e as métricas possíveis de serem calculadas foram descobertas, o *SBINet* já dispõe de todos os campos, fórmulas, funções e algoritmos necessários para a montagem da rede e cálculo das métricas. Realizando a leitura dos datasets, o sistema constrói esta rede internamente e a enriquece atribuindo métodos e propriedades a ela. Por exemplo, por meio da herança de orientação à objetos é possível disponibilizar um método para que cada nó retorne o número de conexões que ele possui. Ao percorrermos todos os nós realizando chamadas ao método e calculando a média do número destas interligações podemos calcular o $coeficiente de agrupamento da rede$, que representa para o usuário a *média de interligações entre estações de bicicletas*. A arquitetura do *SBINet* permite ainda que os demais algoritmos contidos na sua biblioteca de métricas de redes complexas sejam executados sobre os nós e arestas desta rede. É desta forma que todas as métricas descobertas são calculadas e os resultados dos cálculos são acrescentados aos datasets para sejam utilizados posteriormente pelos objetos interativos contidos nos painéis de controle.

Neste momento é gerado um modelo conceitual em baixo nível dos painéis de controle contendo a representação visual de todos os objetos interativos que serão criados. É dada ao usuário a permissão de customizar a ordem de todos os objetos interativos e os seus atributos. A Fig. 6 ilustra este recurso, nela é possível observar que o objeto interativo gráfico de barras está selecionado. Ele será utilizado para exibir os valores da média de interligações entre as estações, descrita no capítulo anterior. Neste momento o sistema preenche os campos contidos na área de customização (à direita da imagem) com os dados do gráfico, onde o usuário pode alterar os atributos de títulos e especificação das dimensões e medidas. A área de ordenação de objetos interativos (ao centro) permite alterar a disposição dos indicadores na tela e a área de criação dos novos objetos interativos permite a inclusão de diversos novos objetos definidos pelo usuário, onde todos os campos dos datasets ficam disponíveis para uso na especificação das medidas e dimensões dos objetos interativos.

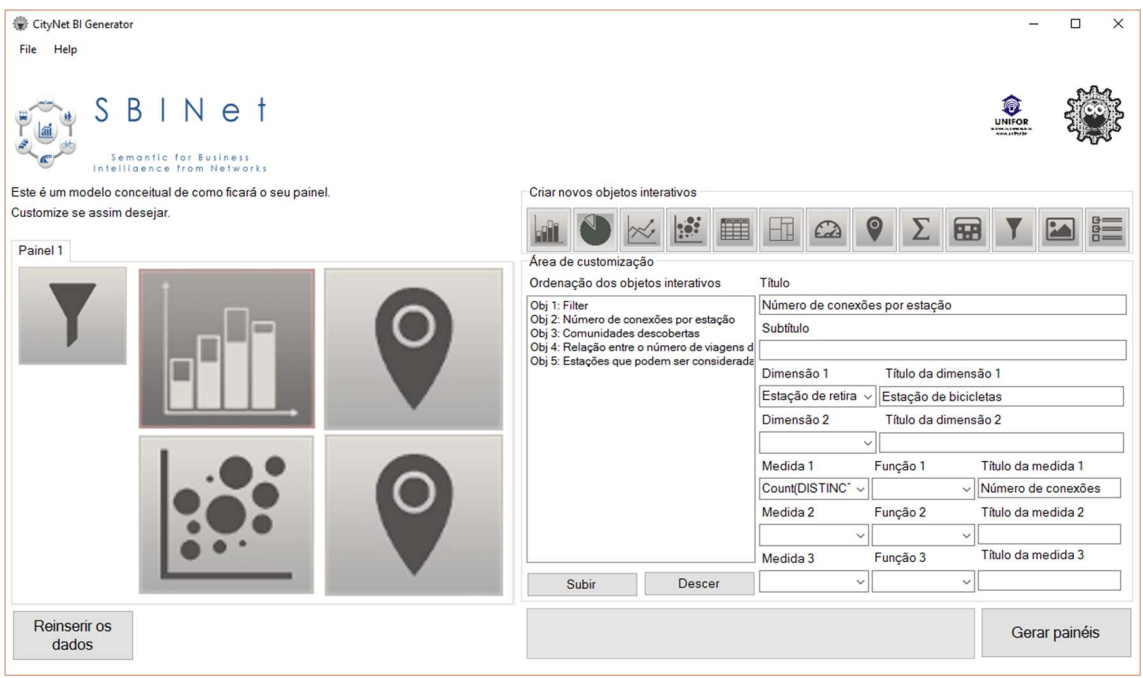

**Figura 6. Modelo conceitual do painél de controle que será gerado**

Finalizado o processo de customização, o *SBINet* está pronto para construir os painéis de controle. Inicialmente o *SBINet* cria aplicações compatíveis apenas com o software de BI Qlik Sense. O início da criação dos painéis se dá pela ação do botão Gerar painéis (Fig. 6), que cria um painel de controle vazio e inicia o processo de preenchimento dele com os objetos interativos associados à cada uma das métricas descobertas. Um a um os gráficos são criados por meio da API de Desenvolvimento do BI utilizado, em nosso exemplo utilizamos a API Qlik Sense .Net SDK 3.2.0. Disponível em https://www.nuget.org/packages/QlikSense.NetSDK/. Finalizado o processo de inserção dos objetos interativos e dos dados contidos nos datasets a aplicação está finalizada e disponível para o uso. O arquivo gerado é depositado diretamente na pasta de aplicações do sistema, permitindo o seu acesso imediato ao abrir o Qlik Sense.

## 5. Analisando métricas e indicadores de sistemas de mobilidade urbana da cidade de Fortaleza

A De posse dos datasets devidamente anotados, realizamos todo o processo de construção dos painéis de controle descritos no capítulo anterior. A Figura 6 ilustra o resultado deste processo. Nela é possível visualizar um dos painéis de controles gerados pelo SBINet por meio dos dados do sistema de bicicletas compartilhadas citado anteriormente aberto pelo Qlik Sense. O gráfico de barras apresenta o número de conexões por estação de bicicleta descrito nos exemplos das ontologias, no processo de descoberta das métricas e nas ações de customização dos objetos interativos. Ele inclui uma linha de referência com a média de conexões por estação de todo o sistema, permitindo identificar as estações que estão dentro da média, quais estão acima e quais estão abaixo. O cálculo deste indicador foi realizado pela métrica de grau médio de redes complexas descrito no capítulo 2. Na mesma Figura é possível visualizar um gráfico de dispersão que cruza o número de conexões das estações com as quantidades de viagens realizadas por elas, o que permite ao usuário analisar a relação entre estas duas métricas. As métricas foram calculadas pelo grau dos nós e pesos das arestas da rede respectivamente. No mapa superior direito são plotadas as estações de bicicletas com uma codificação de cores por comunidades detectadas, neste caso o SBINet detectou 5 comunidades utilizando a métrica de modularidade de redes complexas. No mapa inferior esquerdo é possível identificar em verde as estações com maior probabilidade de serem consideradas como centros do sistema de bicicletas compartilhadas, para isso o software utiliza a métrica de centralidade de grau.

   O gráfico de barras do canto inferior direito apresenta a relação das paradas com os menores volumes de oferta, o que poderiam sugerir as paradas de ônibus candidatas para exclusão da rede. A métrica utilizada neste caso foi a de grau de cara uma das paradas. O mapa no canto superior exibe a maior rota que um usuário conseguiria realizar com apenas uma viagem, de posse dela é possível mensurar o custo máximo de uma viagem dentro da rede, ela foi calculada utilizando a métrica de diâmetro da rede. No canto inferior o mapa exibe em verde as paradas de ônibus que poderiam ser candidatas à terminais de ônibus, considerando a métrica de centralidade de betweenness. Os painéis gerados possuem também filtros, que permitem o usuário a interação com os gráficos e o cruzamento de seleções com o objetivo de realizar uma análise mais aprofundada sobre os dados apresentados.

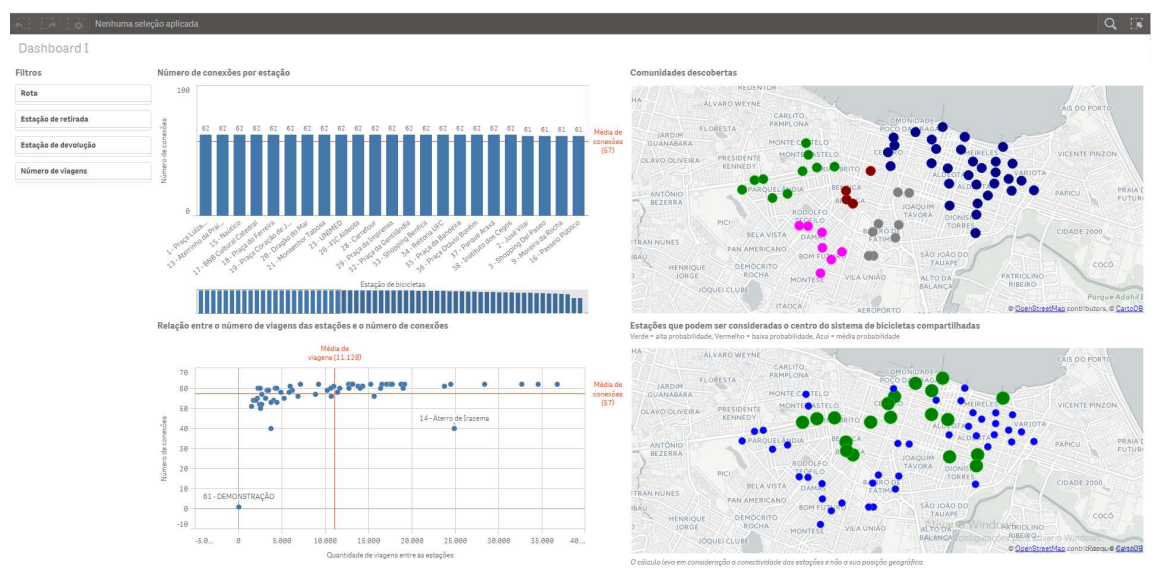

**Figura 7. Painel de controle criado pelo *SBINet* por meio dos dados do sistema de bicicletas compartilhadas da cidade de Fortaleza-CE**

O segundo sistema trata da rede de oferta de ônibus da mesma cidade. Nesta rede, os vértices representam paradas de ônibus e uma aresta que conecta dois vértices $v_i$ e $v_j$, representam a passagem de uma ou mais linhas de ônibus em sequência pelas paradas de ônibus $v_i$ e $v_j$. Aplicamos os dados ao *SBINet*, realizando todo o processo de construção automática dos painéis de controle. A Fig. 8 ilustra o resultado deste processo. Nela é possível visualizar o gráfico de barras no canto superior direito que apresenta as maiores rotas que um usuário conseguiria realizar dentro do sistema utilizando caminhos mínimos, calculadas a partir da métrica de caminhos mínimos de redes complexas. Estas rotas seriam boas candidatas à serem linhas expressas. O gráfico de barras do canto inferior direito apresenta a relação das paradas com os menores volumes de oferta, o que poderiam sugerir as paradas de ônibus candidatas para exclusão da rede. A métrica utilizada neste caso foi a de grau de cara uma das paradas. O mapa no canto superior exibe a maior rota que um usuário conseguiria realizar com apenas uma viagem, de posse dela é possível mensurar o custo máximo de uma viagem dentro da rede, ela foi calculada utilizando a métrica de diâmetro da rede. No canto inferior o mapa exibe em verde as paradas de ônibus que poderiam ser candidatas à terminais de ônibus, considerando a métrica de centralidade de betweenness.

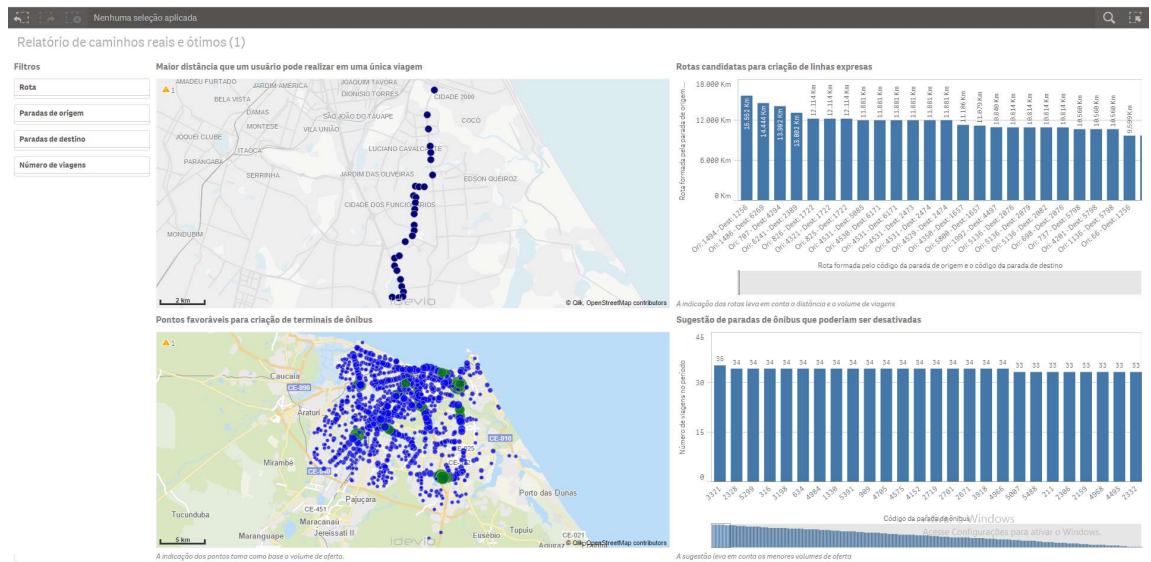

**Figura 8. Painel de controle criado pelo *SBINet* por meio dos dados do sistema de ônibus da cidade de Fortaleza-CE**

Os painéis gerados possuem também filtros, que permitem o usuário a interação com os gráficos e o cruzamento de seleções com o objetivo de realizar uma análise mais aprofundada sobre os dados apresentados.

## 6. Conclusão e trabalhos futuros

Apresentamos uma ferramenta capaz de construir painéis de controle de forma automatizada, tomando como base um conjunto de datasets modelados em forma de uma rede e associados à ontologias que permitam identificar o domínio dos dados e quais métricas melhor se associam ao seu contexto. Os experimentos com sistemas de mobilidade urbana demonstraram a viabilidade técnica e uma significativa abstração dos conhecimentos específicos e necessários de métricas de redes complexas e dos métodos de desenvolvimento de painéis de controle em softwares de BI na criação de indicadores de mobilidade urbana. Acreditamos que as etapas dos processos de construção de uma rede complexa, cálculo das métricas e criação de objetos de interação são facilitadas e podem ser feitas por usuários que não necessariamente conhecem de redes complexas e de construção de aplicações de BI.

A pesquisa ainda pode amadurecer, estamos trabalhando na ampliação das ferramentas de BI utilizadas, com o objetivo de permitir que o usuário escolha em qual software deseja visualizar os dashboards gerados. Pretendemos ampliar o escopo dos domínios cobertos pela aplicação com o objetivo de contemplar outras dimensões além da mobilidade urbana. Por fim há a necessidade de construção de um conjunto de ontologias que contemple estes e outros domínios.